\documentclass[twocolumn]{article}

\usepackage{PRIMEarxiv}

\usepackage[utf8]{inputenc} 
\usepackage[T1]{fontenc}    
\usepackage{url}            
\usepackage{booktabs, multirow} 
\usepackage{amsmath,amsfonts}
\usepackage{nicefrac}       
\usepackage{microtype}      
\usepackage{fancyhdr}       
\usepackage{graphicx}       
\usepackage{array}
\usepackage{textcomp}
\usepackage{stfloats}
\usepackage{verbatim}
\usepackage[caption=false,font=normalsize,labelfont=sf,textfont=sf]{subfig}
\makeatletter 
\def\p@subfigure#1{\thefigure(#1)}
\makeatother
\usepackage{amsthm}
\usepackage{algorithm,algpseudocode}
\usepackage{tabularx}
\usepackage{soul} 
\usepackage{wrapfig} 
\usepackage[numbers,sort&compress]{natbib}
\usepackage[normalem]{ulem}
\useunder{\uline}{\ul}{}
\usepackage[table,xcdraw]{xcolor}

\AtBeginDocument{
  \newgeometry{
    textheight=9in,
    textwidth=7in,
    top=1in,
    headheight=14pt,
    headsep=25pt,
    footskip=30pt
  }
  \setlength{\columnsep}{0.25in}
}
\makeatletter
\renewcommand{\thesubsubsection}{%
  \ifnum\c@subsection=0 \thesection.\arabic{subsubsection}%
  \else\thesubsection.\arabic{subsubsection}\fi}
\makeatother

\newcommand\Tstrut{\rule{0pt}{2.6ex}}         
\newcommand\tstrut{\rule{0pt}{2.3ex}}         
\newcommand\Bstrut{\rule[-0.9ex]{0pt}{0pt}}   

\theoremstyle{plain}
\newtheorem{theorem}{Theorem}[section]

\theoremstyle{definition}
\newtheorem{definition}[theorem]{Definition}

\theoremstyle{remark}
\newtheorem{remark}[theorem]{Remark}

\newcommand{\redit}[1]{\textcolor{black}{#1}}

\newcommand{\noc}{\mathcal{Y}_L}
\newcommand{\oc}{\mathcal{Y}_S}
\newcommand{\tc}{\mathcal{Y}_T}
\newcommand{\low}{\emph{Low}}
\newcommand{\high}{\emph{High}}
\newcommand{\mymethod}{\texttt{FOND}}
\newcommand{\variantF}{\texttt{\mymethod$\backslash$F}}
\newcommand{\variantFB}{\texttt{\mymethod$\backslash$FB}}
\newcommand{\variantFBA}{\texttt{\mymethod$\backslash$FBA}}
\newcommand{\MYmethod}{FOND}

\usepackage{hyperref}       

\makeatletter
\def\bstctlcite{\@ifnextchar[{\@bstctlcite}{\@bstctlcite[@auxout]}}
\def\@bstctlcite[#1]#2{\@bsphack
  \@for\@citeb:=#2\do{%
    \edef\@citeb{\expandafter\@firstofone\@citeb}%
    \if@filesw\immediate\write\csname #1\endcsname{\string\citation{\@citeb}}\fi}%
  \@esphack}
\makeatother

\title{Bringing Equity to Classification: \\Domain Generalization for Domain-Linked Classes}

\author{
  Kimathi Kaai \\
  Dept. of Systems Design Engineering \\
  University of Waterloo \\
  Waterloo, ON, Canada \\
  \texttt{kkaai@uwaterloo.ca} \\
  \And
  Saad Hossain \\
  Dept. of Systems Design Engineering \\
  University of Waterloo \\
  Waterloo, ON, Canada \\
  \texttt{s42hossain@uwaterloo.ca} \\
  \And
  Sirisha Rambhatla \\
  Dept. of Management Sci. \& Eng. \\
  University of Waterloo \\
  Waterloo, ON, Canada \\
  \texttt{sirisha.rambhatla@uwaterloo.ca} \\
}

\begin{document}
\bstctlcite{IEEEexample:BSTcontrol}

\twocolumn[{\maketitle}]

{\renewcommand{\thefootnote}{}\footnotetext{%
  This is the author's accepted version of: K.~Kaai, S.~Hossain, and S.~Rambhatla, ``Bringing Equity to Classification: Domain Generalization for Domain-Linked Classes,'' \textit{IEEE Transactions on Pattern Analysis and Machine Intelligence}, vol.~47, no.~11, pp.~9894--9905, Nov.~2025. DOI: \href{https://doi.org/10.1109/TPAMI.2025.3593407}{10.1109/TPAMI.2025.3593407}. An earlier version appeared in the NeurIPS 2023 Workshop on Distribution Shifts (non-archival). Code: \url{https://github.com/criticalml-uw/fond}}}

\begin{abstract}
Domain generalization (DG) focuses on transferring domain-invariant knowledge from multiple source (training) domains to an \emph{a priori} unseen target domain(s). This task implicitly requires that classes of interest are expressed in multiple sources (\emph{domain-shared}) to break spurious domain-class correlations. However, real-world data scarcity challenges may often result in classes present in only a specific domain (\emph{domain-linked}), which we show leads to extremely poor generalization. In this work, we introduce the domain-linked DG task to the community and develop a methodology to learn useful domain-invariant representations from domain-shared classes for domain-linked ones. Specifically, we propose {\mymethod}, a Fairness-inspired and cONtrastive learning objective for Domain-linked DG. Rigorous and reproducible experimental results communicate that {\mymethod} accomplishes state-of-the-art improvements for domain-linked classes, given a sufficient number of domain-shared classes and with minimal performance trade-offs. Complementary to these contributions, we theoretically analyze this task and provide practical insights for domain-linked class generalizability.
\end{abstract}
\keywords{Domain generalization \and contrastive learning \and image classification}

\section{Introduction} \label{sec:introduction}
Machine learning (ML) has seen significant success across various fields such as computer vision, natural language processing, and healthcare. The primary objective is creating a model capable of learning general and predictive insights from training data, which can then be applied to new (test) data. Traditional ML models are typically trained assuming that training and testing data are identically and independently distributed \textit{i.i.d.} \cite{Vapnik2000TheNO}. However, this assumption often does not hold in real-world scenarios. There are often variations (i.e. \textit{domain-shifts}) between training and testing data, leading to severe degradation in predictive performance \cite{dgsurvey, dgTPAMIsurvey}. 
Gathering data with enough potential domain shifts (lighting, imaging equipment, etc.) to train ML models, is costly and sometimes impossible. Consequently, improving the generalization capability of ML models is crucial in both industry and academic research.

The research field of \textit{domain generalization} (DG) aims to tackle these challenges by developing machine learning techniques to learn discriminative representations that can generalize to data distributions (\emph{domains}) different from those observed during training, i.e. \emph{out-of-distribution} (OOD). In other words, the domain generalization task relaxes the \textit{i.i.d} assumption by allowing source domain distribution(s) at the train time to be different from the target domain (test data). Given this goal, the guiding principle in modern DG algorithms is to learn representations that are invariant to source (training) domains and thus generalizable to unseen target (test) domains \cite{Ye2021TowardsAT, BenDavid2006AnalysisOR}. As a result, recent works aim to explicitly reduce the representation discrepancy between multiple source-domains, by leveraging distribution-alignment \cite{fishr, DIRT2a271795}, adversarial networks \cite{kim2023domain_fourier, Transfer}, domain-based feature-alignment \cite{wang2023pgrad, CAD, Kim2021SelfRegSC}, meta-learning, and few-shot approaches \cite{Qin_2023_CVPR, Gu2022TacklingLC, opendg, ARM, MLDG}. 

\begin{figure}[t]
  \centering
  \includegraphics[width=0.45\textwidth]{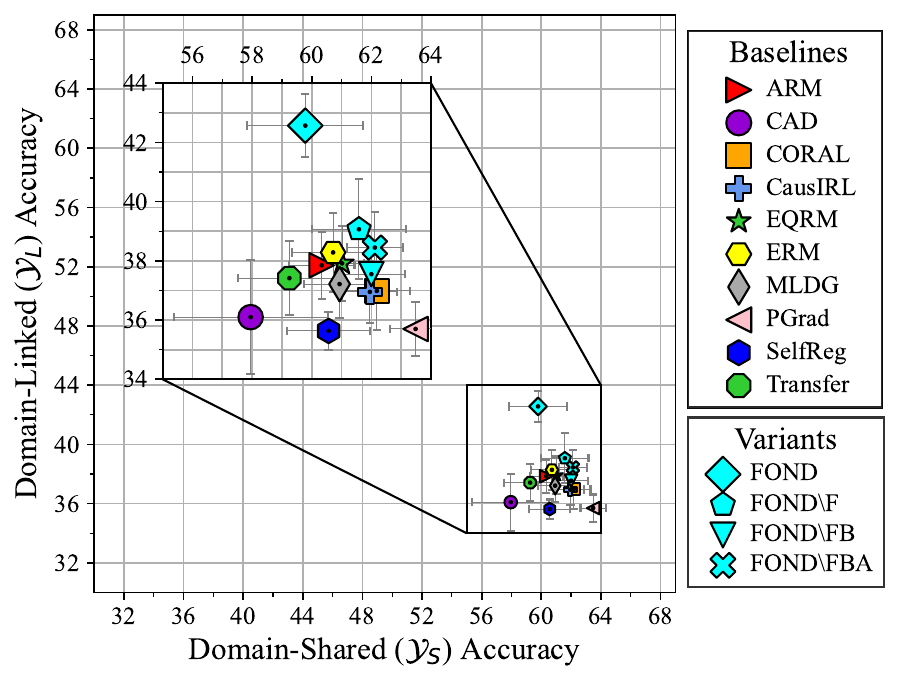}
  \caption{Illustrating domain-linked ($\noc$) and domain-shared ($\oc$) classes performance discrepancies.}
  \label{fig:difference_plot}
\end{figure}

\begin{figure}[t]
    \centering
    \includegraphics[width=0.4\textwidth, trim={1.25cm 0 0 0},clip]{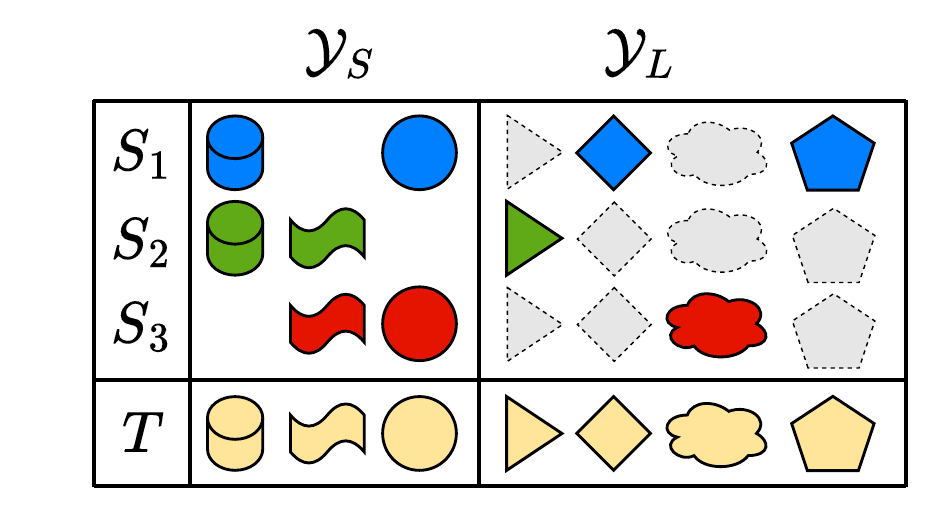}
    \caption{\textbf{Illustrating domain-linked ($\noc$) and domain-shared ($\oc$) classes.} This figure illustrates a shape classification task where the domains are represented by color. During training, some classes are expressed in multiple domains (e.g. circle) while others are expressed in only one domain (e.g. triangle).}
    \label{fig:dl_ds_illustration}
\end{figure}

Despite the significant progress made in DG, existing methods rely on classes being observed in multiple source (training) domains and/or focus only on the overall accuracy. In the real world however, classes of interest may often be observed in specific domains (\emph{domain-linked}, $\noc$), setting them apart from those observed in multiple domains (\emph{domain-shared}, $\oc$); refer to  Fig.~\ref{fig:dl_ds_illustration} for a visual demonstration of these two terms. This leads to generalization challenges in applications such as quality control in manufacturing and food inspection, where certain defects (classes) are often unique/linked to particular types of products (domains) \cite{mvtec}. Similar observations are made in healthcare \cite{ethicalMLhealthcare}, autonomous driving \cite{Piva_2023_WACV}, and fraud detection \cite{fraud} applications where classes of interest may \emph{only} be tied to particular demographics, regions, etc. 

\redit{To illustrate this challenge concretely, we evaluate models on the WILDS\cite{koh2021wilds} Camelyon17 \cite{wilds-camelyon17} dataset, a realistic and clinically relevant histopathology benchmark. In our domain-linked scenario, tumour samples are available exclusively from one hospital domain, reflecting real-world limitations where novel or rare medical conditions initially present at limited locations. Our analysis (see Section~\ref{sec:results-camelyon17}) confirms that existing DG methodologies suffer severe performance degradation when encountering such domain-linked classes.}

Challenges with domain-linked class generalization arise because models tend to exhibit extreme bias towards spurious domain-specific patterns in the absence of observations across domains \cite{lynch2023spawrious, zhang2022correct}. As shown in Fig.~\ref{fig:difference_plot}, the performance discrepancies between domain-linked and domain-shared classes can be extremely severe. Consequently, we seek to improve the generalizability of domain-linked classes while aiming to minimize the domain-shared performance trade-offs. This task, to the best of our knowledge, has not been sufficiently studied.

Motivated by the success of pre-training with classes/objectives different from downstream tasks \cite{he2022masked}, we ask the question -- \emph{can we transfer useful representations from domain-shared classes to domain-linked classes?} We answer this in the affirmative, identifying the cases where it makes sense to utilize the representations learned by domain-shared classes to boost the performance of the domain-linked ones. We first analyze the factors impacting domain-linked class performance theoretically. Leveraging these insights, we motivate and develop a contrastive learning and fairness-inspired objective to disentangle spurious correlations between domain and class, i.e., {\mymethod}, a \texttt{F}airness-inspired and c\texttt{ON}trastive \texttt{D}omain-linked learning strategy.

The goal of our fairness-inspired objective is different from the classic aim of \textit{fair} learning in DG literature, which aims to achieve similar predictive outcomes across protected attributes (gender, race, etc.) \cite{pham2023fairness, MAKHLOUF2021102642, wang2020towards}. On the contrary, our use of fair learning ensures that domain-invariant representations learned from domain-shared classes still benefit domain-linked ones. To complement this goal, we focus the model's learning on specific pairwise relationships via a specialized contrastive learning objective introduced in Section~\ref{sec:methodology}.

We rigorously evaluate {\mymethod} and its variants on different types of domain-shifts, domain-shared class distributions, and total number classes. To this end, our experiments summarize over 360 trials for each of our eleven DG baselines, including SOTA \cite{urm, rdm}, on three established DG benchmark datasets (PACS \cite{Li2017DeeperBA_pacs}, VLCS \cite{Fang2013UnbiasedML_vlcs}, OfficeHome \cite{OfficeHome}) and the clinically impactful WILDS \cite{koh2021wilds} Camelyon17 \cite{wilds-camelyon17} medical imaging dataset. We find that {\mymethod} improves domain-linked class performance when observing a high enough number of domain-shared classes. Interestingly, we observe that the classic \texttt{ERM} is still a strong baseline, but {\mymethod} and its variants achieve a remarkable overall performance improvement of $+9.3$ over \texttt{ERM} on average ($26.9\%$ improvement), with a gain of $39.2\%$ on VLCS. \redit{On Camelyon17, {\mymethod} surpasses the best-performing baseline by nearly 9\% in tumor detection accuracy on out-of-distribution test data.} We also compare our domain-shared performance with SOTA, showing that domain-linked performance can be significantly boosted with a modest domain-shared class performance trade-off.

\subsection{Contributions}
\noindent We summarize our contributions as follows:
\begin{enumerate}
    \item Identifying the severe performance discrepancy between domain-linked and domain-shared class performance in existing DG methods.
    \item Proposing $\mymethod$; the first method for improving domain-linked class generalization using representations learned from domain-shared classes.
    \item Leveraging existing domain adaptation theory to analyze when and why domain-shared classes usually generalize better than domain-linked ones.
    \item Accomplishing state-of-the-art performance for domain-linked DG while providing practical insights into the data scarcity conditions that affect their generalizability. 
\end{enumerate}

\section{Related Works} \label{sec:related-works}


\subsubsection{Data Manipulation} 
\noindent These techniques primarily focus on data augmentation and generation techniques such as affine transformations in conjunction with additive noise, cropping and so on \cite{Shorten2019ASO, he2016deep}. Other methods include simulations \cite{Tobin2017DomainRF, Yue2019DomainRA, Tremblay2018TrainingDN}, gradient-based perturbations like CrossGrad \cite{Shankar2018GeneralizingAD}, adversarial augmentation \cite{Volpi2018GeneralizingTU} and image mixing (e.g. CutMix \cite{Mancini2020TowardsRU}, Mixup \cite{zhang2018mixup} and Dir-mixup \cite{opendg}). Furthermore, generative models using VAEs GANs are also popular techniques for diverse data generation \cite{Anoosheh2017ComboGANUS, Zhou2020LearningTG, Somavarapu2020FrustratinglySD, Huang2017ArbitraryST}. Since a model's generalizability is a function of training data diversity \cite{Vapnik2000TheNO}, {\mymethod} and other approaches should be used in conjunction; therefore, this paper's analysis does not include these methodologies. 

\subsubsection{Multi-Domain Feature Alignment} 
\noindent Most existing DG approaches focus on minimizing differences between features learned from multiple source domains, aiming to create representations invariant to domain shifts and robust to unseen target domains. Techniques include aligning cross-domain features using statistical distance metrics like mean and variance (e.g., CORAL \cite{CORAL}, M$^3$SDA \cite{Peng2019MomentMF}), and measures such as Wasserstein distance and KL Divergence \cite{zhou2021domain, Wang2021ClassconditionedDG}. Methods like DIRT \cite{DIRT2a271795} and MDA \cite{Hu2019DomainGV} transform feature distributions across domains before minimizing discrepancies, while others like PGrad \cite{wang2023pgrad} and Fishr \cite{fishr} apply techniques to optimization gradients. Recent works, such as CausIRL \cite{causirl}, use structural causal models for domain-invariant representations, while EQRM \cite{eastwood2022probableEQRM} criticizes average performance optimization and proposes a probabilistic framework. These approaches, relying on observing multiple domain shifts per class, often degrade the generalization performance of domain-linked classes.

\subsubsection{Domain-Discriminative Adversarial Training}
\noindent Different from explicitly aligning features across multiple sources, adversarial training formulates domain invariant features learning as an adversarial game \cite{ZhuADV_2022_CVPR, yangADV2021, shaoADV2019multi, Li2018DomainGW, Gong2018DLOWDF}. Simply speaking, the learning objective is to stop a discriminator from identifying the source domain of a set of learned features. However, since there is a correlation between $\noc$ classes and their domains, adversarial domain discriminators may infer a samples domain from class-discriminative features.

\subsubsection{Meta-Learning} 
\noindent The rationale for applying meta-learning to domain generalization is to emulate domain shifts during training, hoping that the model can better handle domain shifts in unseen domains. These approaches divide the source domains into non-overlapping meta-train and meta-test episodes. The popular MAML \cite{maml} architecture inspires most DG meta-learning approaches, most notably the foundational MLDG \cite{MLDG} and ARM \cite{ARM} methods that have become popular base DG architectures \cite{zhong2022metaMOE, opendg}. However, meta-learners still suffer challenges when classes cannot be observed in both the meta-train and meta-test episodes. Consequently, approaches like Transfer \cite{Transfer} combine meta-learning with adversarial techniques to identify more generalizable representations.

\subsubsection{Contrastive Learning} 
\noindent Contrastive learning aims to learn representations, self-supervised \cite{Chen2020ASF} or supervised \cite{Khosla2020SupervisedCL, Motiian2017UnifiedDS}, such that similar samples are embedded close to each other while distancing dissimilar samples \cite{Huang2020RCSSelfChallengingIC, CAD, Kim2021SelfRegSC, Khosla2020SupervisedCL, Chen2020ASF, Motiian2017UnifiedDS}. These methods focus on multi-domain comparisons to identify domain-invariant representations; see \cite{dgTPAMIsurvey} and the references therein. However, this is insufficient for identifying generalizable representations for domain-linked classes.

\subsubsection{Fairness} 
\noindent Notions of fairness in DG \cite{MAKHLOUF2021102642} involves reducing the performance discrepancies (biases) between protected attributes (e.g. demographic) \cite{pham2023fairness, wang2020towards}. Motivated by these works, we impose a fairness-inspired learning objective to learn generalizable representations for domain-linked classes.

\subsubsection{Label Shift in Generalization}
\noindent Our focus on domain-linked class generalization falls under the broad and recognized concept of \textit{label-shift} in DG literature. Label shift occurs when source/target domains have different class distributions. This, however, is too broad of a categorization; therefore, published DG methodologies primarily focus on one of two types: 1) Works like \cite{liu2021domain, causirl, Transfer} assume distribution differences in training domains while maintaining that all classes are present in each source domain; 2) Works like \cite{Chen_2023_ICCV, liu2022unknown} focus on identifying unknown (novel) target-domain classes, i.e., \textit{open-set generalization}. We uniquely focus on improving the performances of classes present in only one domain while maintaining that all target classes are observed during training. Furthermore, our baselines were selected to cover a variety of DG methodologies that expect (implicitly and explicitly) differences in label distributions.

\section{Problem Formulation} \label{sec:problem-formulation}

\noindent We formalize some definitions for the purposes of the paper. First, we define a domain as follows. 
\begin{definition}[Domain]
\label{def:domain}
    \emph{Let $\mathcal{X}$ denote a nonempty input space (e.g. images, text, etc) and $\mathcal{Y}$ an output label space. We denote as specific domain as $S = \{(\mathbf{x}_j, y_j)\}_{j=1}^{n} \sim \mathcal{D}^{S}: \mathcal{X}^{S} \times \mathcal{Y}^{S}$, where $\mathbf{x} \in \mathcal{X}^{S} \subseteq \mathbb{R}^d$ and $y \in \mathcal{Y}^{S} \subset \mathbb{Z}$.}
\end{definition}
Given this domain definition, the DG task -- which entails learning representations from multiple source domains to generalize to unseen target domain (s) -- can be formalized as shown below.
\begin{definition}[Domain generalization]
\label{def:dg}
    \emph{Given $K$ training (source) domains $\mathcal{S} = \{ S^i~|~i = 1,..., K \}$ where $\textstyle S^{i} = \{(\mathbf{x}^i_j, y^i_j)\}_{j=1}^{n_i}$ denotes the $i$-th source domain with $n_i$ samples, and the joint distributions between each pair of domains are different: $\mathcal{D}^{S^i} \neq \mathcal{D}^{S^j} : 1 \leq i \neq j \leq K$. Then the goal is to learn a predictive function from $\mathcal{S}$ for reliable performance on an unseen, out-of-distribution target-domain $T \sim \mathcal{D}^{T}: \mathcal{X}^T \times \mathcal{Y}^T$ (i.e. $\mathcal{D}^{T} \neq \mathcal{D}^{S^i}$ for $i\in \{1,...,K\}$).}
\end{definition}

We evaluated methods for \emph{closed-set} domain generalization (i.e. $\mathcal{Y}^T = \bigcup^K_{i=1}\mathcal{Y}^{S^i}$) where no single source domain expresses all target classes (i.e., $\mathcal{Y}^{T} \subset \mathcal{Y}^{S^i}$ for $i\in \{1,...,K\}$). Furthermore, during training, there exists a set of classes expressed in only one source domain, i.e. \emph{domain-linked} classes $\mathcal{Y}^{S}_{\lambda}$.

\begin{definition}[Set of domain-linked classes]
\label{def:dl}
    $\mathcal{Y}^{S}_{\lambda} = \{y: |\{S \in \mathcal{S}, y \in \mathcal{Y}^{S} \}|=1 \}$, $\noc$ for brevity.
\end{definition}
Similarly, classes expressed in multiple domains, i.e. \emph{domain-shared} classes $\mathcal{Y}^{S^{i}}_{\sigma}$ as shown below. Here, $\mathcal{Y}^T = \noc \cup \oc$ and $\noc \cap \oc = \emptyset $.
\begin{definition}[Set of domain-shared classes]
\label{def:ds}
    $\mathcal{Y}^{S}_{\sigma} = \{y: |\{S \in \mathcal{S}, y \in \mathcal{Y}^{S} \}|>1 \}$,  $\oc$ for brevity.
\end{definition}

\subsection{Learning Objective}
\noindent The learning objective is to identify a generalizable predictive function $M: \mathcal{X} \rightarrow \mathcal{Y}$ to achieve a minimum predictive error on an unseen, out-of-distribution, test domain $T$ under the previously outlined conditions. The task network is defined as $M=(F\circ G)(\mathbf{x})$, to learn the feature extractor $F: \mathcal{X} \rightarrow \mathcal{H}$ and the classifier $G: \mathcal{H} \rightarrow \mathcal{Y}^\Delta$, where $\Delta$ is a probability simplex over the set of classes $\mathcal{Y}^T$.

\section{Theoretical Analysis} \label{sec:theory}

\begin{algorithm*}[t]
\caption{{\mymethod} Training Algorithm}
\label{alg:learning_algorithm}
\small

    \begin{algorithmic}
    \Require Source datasets $\mathcal{S}$; Feature extractor $F$; Projection network $Q$; Classification network $G$; Inter-domain positive weight $\alpha$; Intra-domain negative weight $\beta$
    \While{Not Converged}
        \State Sample a batch of data $\mathcal{B}=\{(\mathbf{x}^1,\mathbf{y}^1), (\mathbf{x}^2,\mathbf{y}^2), ..., (\mathbf{x}^K,\mathbf{y}^K)\}$ from all source domains $\mathcal{S}$
        \State $\mathcal{B}_F = \{(\mathbf{h}^1,\mathbf{y}^1), (\mathbf{h}^2,\mathbf{y}^2), ..., (\mathbf{h}^K,\mathbf{y}^K)\} \gets F(\mathcal{B}) $ \Comment{Generate the intermediate representations}
        \State $\mathcal{B}_Q = \{(\mathbf{z}^1,\mathbf{y}^1), (\mathbf{z}^2,\mathbf{y}^2), ..., (\mathbf{z}^K,\mathbf{y}^K)\} \gets Q(\mathcal{B}_F) $ \Comment{Generate the feature projections}
        \State $\mathcal{L}_{xdom} \gets \{\mathcal{B}_Q, \alpha, \beta\}$ \Comment{Calculate domain-aware loss according to  \eqref{eq:xdom-loss}}
        \State $\mathcal{B}_C \gets G(\mathcal{B}_F) $ \Comment{Generate the classification logits}
        \State $\mathcal{B}_C^{(L)}, \mathcal{B}_C^{(S)} \gets \{\mathcal{B}_C\}$ \Comment{Separate logits based on ground-truth label group, i.e. $\noc$ or $\oc$}
        \State $\mathcal{L}_{fair}, \mathcal{L}_{task} \gets \{\mathcal{B}_C^{(L)}, \mathcal{B}_C^{(S)}\}, \{\mathcal{B}_C\}$ \Comment{Calculate task and fair losses according to \eqref{eq:fair-loss}}
        
    \EndWhile
    \end{algorithmic}
        \Return $F, G$
        
\end{algorithm*}

\noindent In this section, we theoretically formalize the challenges faced by domain-linked classes by demonstrating the roots of the performance discrepancies between domain-linked and domain-shared classes. The next section uses these insights to develop the proposed algorithm. Our main result comes in the form of the following theorem.

\begin{theorem} \label{thm:main}
Given $K$ source domains ${S^i}$ for $i\in[1,K]$, each contributing $\alpha_i\cdot m$ i.i.d. samples where $\sum_{i=1}^{K} \alpha_i = 1$, and each $\alpha_i$ denotes the fraction of samples contributed by a source $S^i$, a target domain $T$, and a hypothesis class $\mathcal{H}$. The error $\epsilon_{T_\ell}(\hat{h})$ of a domain-linked class $\ell$ of hypothesis $\hat{h}\in \mathcal{H}$, an empirical minimizer of $\epsilon(h)$ on the source samples, is upper bounded by
\begin{equation}
    \label{eq:error_bounds_linked}
    \epsilon_{T_{\ell}}(\hat{h}) \leq \epsilon_{T_{\ell}}(h_T^*)  + d_{\mathcal{H}\Delta\mathcal{H}}(T, S^{\ell}) + \mathcal{O}(m^{-1}), 
\end{equation}
with probability at least $(1-\delta_\ell)$ for $\delta_\ell \in (0,1)$, where $\epsilon_{T_\ell}(h_T^*)$ represents the  error achieved by $h_T^* = {\min}_{h\in\mathcal{H}}~ \epsilon_T(h)$ on the target domain for the class $\ell$. Further, the upper-bound, if this were a domain-shared class expressed in $K$ domains, is
\begin{equation}
    \label{eq:error_bounds_shared}
    \epsilon_{T_{\ell}}(\hat{h}) \leq \epsilon_{T_{\ell}}(h_T^*)  + \sum^K_{i=1} \alpha_i~d_{\mathcal{H}\Delta\mathcal{H}}(T, S^i) + \mathcal{O}(m^{-1}),
\end{equation}
 with probability at least $(1-\delta_s)$ for $\delta_s \in (0,1)$, where $d_{\mathcal{H}\Delta\mathcal{H}}(\cdot, \cdot)$ is the average distribution divergence between domains. Then, if   $d_{\mathcal{H}\Delta\mathcal{H}}(T, S^{i}) \leq d_{\mathcal{H}\Delta\mathcal{H}}(T, S^{\ell})$ for all $i$, the upper-bound in \eqref{eq:error_bounds_linked} is greater than that of \eqref{eq:error_bounds_shared}.
\end{theorem}
\begin{remark} \label{rem:1}
    The result shows the dependence on the number of samples. While we use $\mathcal{O}(m^{-1})$ to simplify this dependence, the number of samples for domain-linked classes is usually lower than those for domain-shared classes, which is a significant source of the performance discrepancy. 
\end{remark}
\begin{remark} \label{rem:2}
    The average distribution divergence is another source of performance discrepancy if the domain-shared classes have a larger and more expressive dataset than domain-linked the condition $d_{\mathcal{H}\Delta\mathcal{H}}(T, S^{i}) \leq d_{\mathcal{H}\Delta\mathcal{H}}(T, S^{\ell})$ is easy to meet. Although not explicitly stated, the diversity of data can be encoded as the span of a given dataset, i.e. the size of the convex hull of all data points.
\end{remark}

This theoretical result demonstrates that the distribution divergence and samples play a major role in a fixed hypothesis class. While the number of samples is usually fixed, we aim to inject domain diversity from shared classes to improve the performance of domain-linked ones. Our empirical result corroborates this theoretical intuition that a high-shared setting, where a higher number of shared classes are observed in different domains, indeed helps domain-linked classes. 

\subsection{Proof Sketch}

\noindent Given $K$ source domains and a class $c\in[C]$, we use the result from Theorem 4 in \cite{ben2010theory} assuming that each of the domains contributes $\alpha_i\cdot m$ i.i.d. samples where $\sum_{i=1}^{K} \alpha_i = 1$ without loss of generality. The following result is reproduced from \cite{ben2010theory}, where we have used the Landau notation to simplify the result. Additionally, for our analysis, $\lambda$-closeness ($\lambda_\alpha = 0$) is assumed because of the extremely large capacities of neural networks; this assumption is often made in DG and is experimentally validated \cite{NEURIPS2021_ecf9902e_empirical}. Furthermore, since $\epsilon_T(h_T^*)$ is fixed, our analysis focuses on the $d_\alpha$ and $O(m^{-1})$ terms.
\begin{theorem}[\cite{ben2010theory} Theorem~4]\label{thm:shai}
Given $K$ source domains ${S^i}$ for $i\in[1,K]$, each contributing $\alpha_i\cdot m$ i.i.d. samples where $\sum_{i=1}^{K} \alpha_i = 1$, and each $\alpha_i$ denotes the fraction of samples contributed by a source $S^i$, a target domain $T$, that the hypothesis class $\mathcal{H}$ needs to generalize to. Then, if $\epsilon_{T_c}(h_T^*)$ represents the  error achieved by hypothesis $h_T^* = \min_{h\in\mathcal{H}}~ \epsilon_T(h)$ on a class $c\in [1,C]$ on the target domain, the error $\epsilon_{T_c}(\hat{h})$ of $c$ on the target domain $T$ using hypothesis $\hat{h}\in \mathcal{H}$ , an empirical minimizer of $\epsilon(h)$ on the source samples is upper bounded by 
\begin{equation}
    \label{eq:error_bounds}
    \epsilon_{T_c}(\hat{h}) \leq \epsilon_{T_c}(h_T^*)  + \sum^K_{i=1} \alpha_i~d_{\mathcal{H}\Delta\mathcal{H}}(T, S^i) + \mathcal{O}(m^{-1}) 
\end{equation}
with probability at least $(1-\delta)$, where $d_{\mathcal{H}\Delta\mathcal{H}}(\cdot, \cdot)$ is the average distribution divergence between domains, and $\delta \in (0,1)$.
\end{theorem}

Now, let $\ell$ and $s$ denote a domain-linked and domain-shared class, respectively. We will use Theorem~\ref{thm:shai} to evaluate the upper bound on error for $\ell$ and $s$.

For class $\ell$, there is just one source domain therefore $\alpha_i = 1$, and we have 
\begin{equation*}
    \epsilon_{T_{\ell}}(\hat{h}) \leq \epsilon_{T_{\ell}}(h_T^*)  + d_{\mathcal{H}\Delta\mathcal{H}}(T, S^{\ell}) + \mathcal{O}(m^{-1}), 
\end{equation*}
with probability at least $(1-\delta_\ell)$ for $\delta_\ell \in (0,1)$, where $\epsilon_{T_\ell}(h_T^*)$ represents the  error achieved by $h_T^* = {\min}_{h\in\mathcal{H}}~ \epsilon_T(h)$ on the target domain for the class $\ell$.

Now, if the class $\ell$ is expressed in $K$ domains, then the upper-bound from Theorem~\ref{thm:shai} is given by 
\begin{equation*}
    \epsilon_{T_{\ell}}(\hat{h}) \leq \epsilon_{T_{\ell}}(h_T^*)  + \sum^K_{i=1} \alpha_i~d_{\mathcal{H}\Delta\mathcal{H}}(T, S^i) + \mathcal{O}(m^{-1}),
\end{equation*}

Analyzing the upper bounds above, we note that for the case when $\ell$ is a domain-shared class if any of the source domains $S^i$ are closer to the target domain $T$ as compared to when it is a domain-linked class, i.e.
\begin{align*}
d_{\mathcal{H}\Delta\mathcal{H}}(T, S^{i}) \leq d_{\mathcal{H}\Delta\mathcal{H}}(T, S^{\ell}) ~\forall~ i,
\end{align*}
the upper-bound for $\epsilon_{T_{\ell}}(\hat{h})$ is greater than $\epsilon_{T_{s}}(\hat{h})$. The bounds are equal when $d_{\mathcal{H}\Delta\mathcal{H}}(T, S^{i}) = d_{\mathcal{H}\Delta\mathcal{H}}(T, S^{\ell})$ for all $i$.


\begin{figure}[t]
    \centering
    \includegraphics[width=0.25\textwidth]{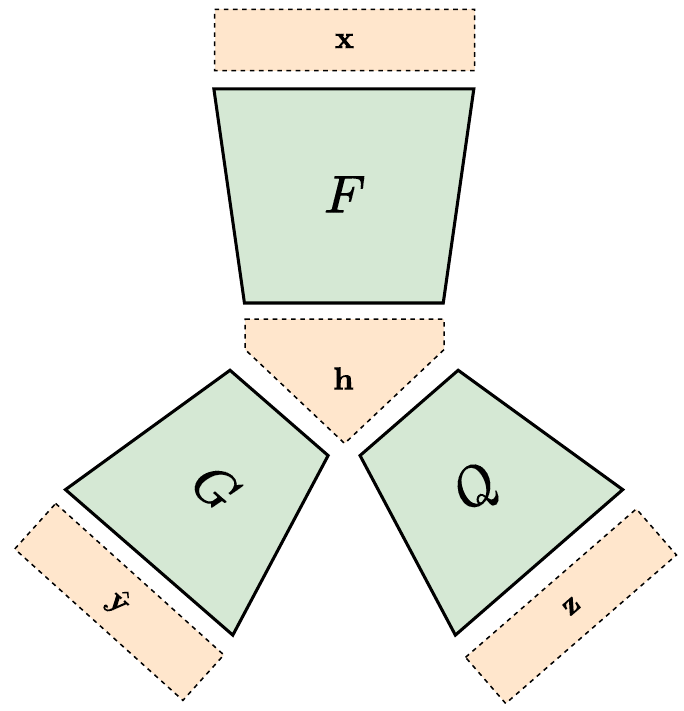}
    \caption{\textbf{Visualizing the {\mymethod} model architecture.} For an input, $\mathbf{x}$, the feature extractor $F(.)$ generates a representation vector, $\mathbf{h}$. The network $G(.)$ performs the downstream classification task, and $Q(.)$ generates the low-dimension representation $\mathbf{z}$ used for the contrastive objective, \eqref{eq:xdom-loss}.}
    \label{fig:architecture}
\end{figure}

\section{Methodology} \label{sec:methodology}


\noindent We introduce the learning algorithm {\mymethod}, which learns domain-invariant representations from domain-shared $\oc$ classes that improve domain-linked $\noc$ class generalization.  We achieve this by minimizing the following objective as detailed in Algorithm~\ref{alg:learning_algorithm}:

\begin{equation}
\label{eq:overall-loss}
    \mathcal{L}_{\mymethod} = \mathcal{L}_{task} + \lambda_{xdom} \cdot \mathcal{L}_{xdom} + \lambda_{disc} \cdot \mathcal{L}_{disc}.
\end{equation}

We impose domain-invariant representation learning by focusing on specific pairwise sample relationships through the contrastive objective $\mathcal{L}_{xdom}$. Since we require these representations to improve domain-linked class generalizability, we impose a fairness-inspired performance discrepancy loss between $\noc$ and $\oc$ classes through the $\mathcal{L}_{disc}$ objective. The following sections describe the formulation of these objectives.

\subsection{Pair-Wise Relationships for Domain-Invariance}

\noindent The $\oc$ and $\noc$ performance discrepancy (Fig. \ref{fig:difference_plot}) results since algorithms do not observe enough domain-variances in $\noc$ data as they do with $\oc$ data. In alignment with Remark~\ref{rem:2}, in Section~\ref{sec:theory}, observing domain variances allows models to learn representations that reduce the average distribution divergence between source and target domain data. Therefore, unlike modern feature-alignment DG approaches, our learning objective must treat $\noc$ and $\oc$ classes differently.

Maximizing the mutual information between positive (same-class) inter-domain samples guides domain-invariant learning \cite{simclr, Ren_2023_CVPR}. For example, in Fig.~\ref{fig:dl_ds_illustration}, an algorithm observing pairwise relationships between samples from domain-shared $\oc$ classes may observe that encoding edge features increases mutual information while color reduces it. Furthermore, due to the success of \textit{hard negative mining} in representation learning literature \cite{robinson2021contrastive, zhang2022m, liu2023hard}, we hypothesize that negative (different-class) intra-domain comparisons are more informative than negative inter-domain comparisons for reducing spurious domain and class correlations. For example, in Fig.~\ref{fig:dl_ds_illustration}, an algorithm may achieve color invariance by minimizing mutual information between samples from different classes (shapes) but from the same domain (color). This is validated in Table~\ref{table:table-results} where we observe that variants including both $\alpha$ and $\beta$ (i.e., \texttt{FOND} and \texttt{FOND$\backslash$F}) outperform all baselines. 

Therefore, we define a feature extractor $F: \mathcal{X} \rightarrow \mathcal{H}$ to take a input samples $\mathbf{x} \in \mathcal{X}$ and generate representation vectors, $\mathbf{h} \in \mathcal{H} \subseteq \mathbb{R}^{d_F}$. We regularize the representation vectors by applying a contrastive objective to the output of a projection network $Q: \mathcal{H} \rightarrow \mathcal{Z}$ that generates normalized, lower-dimensional representations $\mathbf{z} \in \mathcal{Z} \subseteq  \mathbb{R}^{d_Q}$. The goal of the contrastive objective defined by \eqref{eq:xdom-loss} is to maximize the cosine similarity of projected representations $\mathbf{z}$ between same-label samples and minimize those that do not.%
 
{\small
\begin{align}
\label{eq:xdom-loss}
\hspace{-5pt}\mathcal{L}_{xdom} &= \sum_{i \in I} \tfrac{-1}{|P(i)|} \sum_{p \in P(i)} \log ~\frac{\alpha \cdot \exp(\mathbf{z}_i \cdot \mathbf{z}_p / \tau)}{\textstyle\sum_{j \in I \backslash \{i\}} \beta \cdot \exp(\mathbf{z}_i \cdot \mathbf{z}_j / \tau)} \notag \hspace{50pt} & \\
    \alpha(\mathbf{z}_i,\mathbf{z}_p) &= \begin{cases} a, & S{(\mathbf{z}_i)} \neq S{(\mathbf{z}_p)},~\text{where}~a \geq 1 \\ 1, & \text{otherwise}\end{cases} \notag \\
    \beta(\mathbf{z}_i, \mathbf{z}_j) &= \begin{cases} b, & S{(\mathbf{z}_i)} = S{(\mathbf{z}_j)}, y_i \neq y_j,~\text{where}~b \geq 1 \\ 1, & \text{otherwise}\end{cases}
\end{align}}

Let $i \in I \equiv \{1...N\}$ be the index of a sample (denoted as the \emph{anchor}) where $N$ denotes the batch size. $P(i) = \{p \in I \backslash \{i\}: y_p = y_i\}$ is the set of indices of all positives in the batch. The term $\alpha$ increases the cosine similarity weight of the anchor $\mathbf{z}_i$ and positive $\mathbf{z}_p$ sample if they are inter-domain ($S(\mathbf{z}_i) \neq S{(\mathbf{z}_p)}$) pairs. Note that $S{(\mathbf{z}_i)}$ denotes the domain $S \in \mathcal{S}$ that $\mathbf{z}_i$ belongs to. Additionally, $\beta$ increases the cosine-similarity weight of the anchor $\mathbf{z}_i$ and $\mathbf{z}_j$ if they are negative, intra-domain ($S{(\mathbf{z}_i)} = S{(\mathbf{z}_j)}$) pairs. 

Based on the output values for $\alpha$ and $\beta$, {\mymethod} can be represented in multiple ways. The {\variantFBA} method variant sets $a = b = 1$, {\variantFB} sets $a \geq 1, b = 1$ and {\variantF} sets $a \geq 1, b \geq 1$; these variants omit fairness (i.e. \texttt{$\backslash$F}). Note that $a$ and $b$ are hyper-parameters. Table~\ref{table:table-results} reports the performances of {\mymethod}'s and its variants concerning other DG baselines.

\subsection{Transferring Domain-Invariant Representations}

\noindent Increasing the weight of positive inter-domain similarity metrics through $\alpha$ in \eqref{eq:xdom-loss} biases the model towards domain-shared $\oc$ generalization since these metrics are not present between domain-linked $\noc$ class samples. Therefore, we draw inspiration from notions of \textit{fair} learning in DG literature to learn generalizable features for $\noc$ classes.

Notions of fair learning in DG literature aim to makepredictive outcome $\hat{Y}$ of a model $M(X)$ independent of protected attributes $A$ (e.g., gender, race, age) present in a dataset \cite{ pham2023fairness, MAKHLOUF2021102642, wang2020towards}. Directly applying this notion of fair learning by defining our protected attribute $A$ as whether a sample belongs to a domain-shared or domain-linked class would make the prediction outcome $M(X)=\hat{Y}$ completely dependent on $A$.



To overcome these challenges, we constrain the \emph{error rates} across domain-shared and domain-linked classes. The objective is to enforce that the domain-invariant representations learned from domain-shared $\oc$ classes \textit{also} improve the generalizability of the disadvantaged domain-linked $\noc$ classes. The violation of this objective is measured by \eqref{eq:fair-loss}. We observe in Section~\ref{sec:results} that this leads to large improvements in $\noc$ classes, given the model observes a sufficient number of $\oc$ classes; this is denoted as the {\high} shared-class experimental setting defined in Section~\ref{sec:class-settings}. Note that since the task is image classification, the task loss, $\mathcal{L}_{task}$, is categorical cross-entropy.

\begin{equation}
\label{eq:fair-loss}
    \mathcal{L}_{disc} = |\mathcal{L}_{task}^{\noc} - \mathcal{L}_{task}^{\oc}|
\end{equation}

\subsection{Model Architecture} \label{sec:model-architecture}
\noindent We now explain each module of {\mymethod}; refer to Fig.~\ref{fig:architecture}.
The \textit{Feature Extraction Network}, $F(.)$, takes a training input sample $\mathbf{x} \in \mathcal{S}$ and generates a representation vector, $\mathbf{h} = F(\mathbf{x}) \in \mathbb{R}^{d_F}$ where $d_F = 512$. The hidden representations are normalized, $||\mathbf{h}|| = 1$, since this has experimentally been shown to improve performance \cite{Schroff2015FaceNetAU}.
The \textit{Projection Network}, $Q(.)$, takes the representation vector $\mathbf{h}$ and non-linearly projects it to a lower-dimensional vector $\mathbf{z} = P(\mathbf{h}) \in \mathbb{R}^{d_Q}$ where $d_Q=256$. Additionally, the projection vector is normalized $||\mathbf{z}|| = 1$. These projections are used for {\mymethod}'s contrastive learning objective \eqref{eq:xdom-loss}. Previous works \cite{Chen2020ASF, Khosla2020SupervisedCL} validate separating the representations $\mathbf{h}$ used for downstream tasks and those for contrastive objectives $\mathbf{z}$. The network is implemented by a 2-layer MLP with a ReLU activation function between the layers.
The \textit{Classification Network}, $G(.)$, performs the image classification downstream task with the representations generated by $F(.)$, i.e., $\mathbf{h} \in \mathbb{R}^{d_F}$. The network's output is a vector of dimension $|\tc|$ denoting the softmax label probabilities of the input $\mathbf{x}$. A 2-layer MLP implements the network with a ReLU activation function between the layers. 

\section{Setup}\label{sec:setup}

\begin{figure}[t]
    \centering
    \includegraphics[width=\columnwidth]{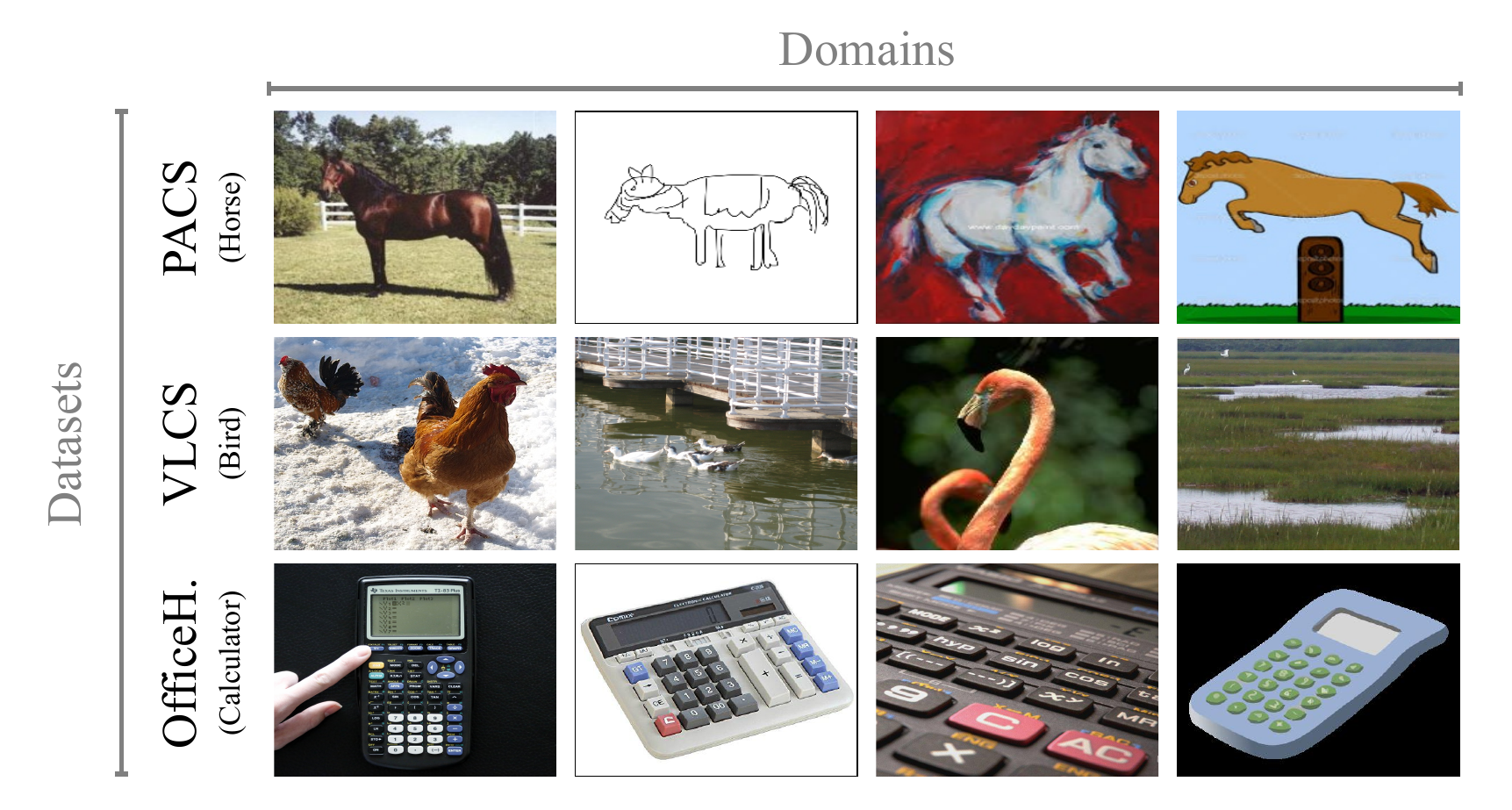}
    \caption{\redit{\textbf{Visualizing the domain shifts of standard multiclass domain generalization datasets.} PACS \cite{Li2017DeeperBA_pacs} and OfficeHome \cite{OfficeHome} exhibit obvious, style-based, domain variations, while VLCS \cite{Fang2013UnbiasedML_vlcs} is more real-world and nuanced.}}
    \label{fig:data-multiclass}
\end{figure}

\begin{figure}[t]
    \vspace{-10pt}
    \centering
    \includegraphics[width=\columnwidth]{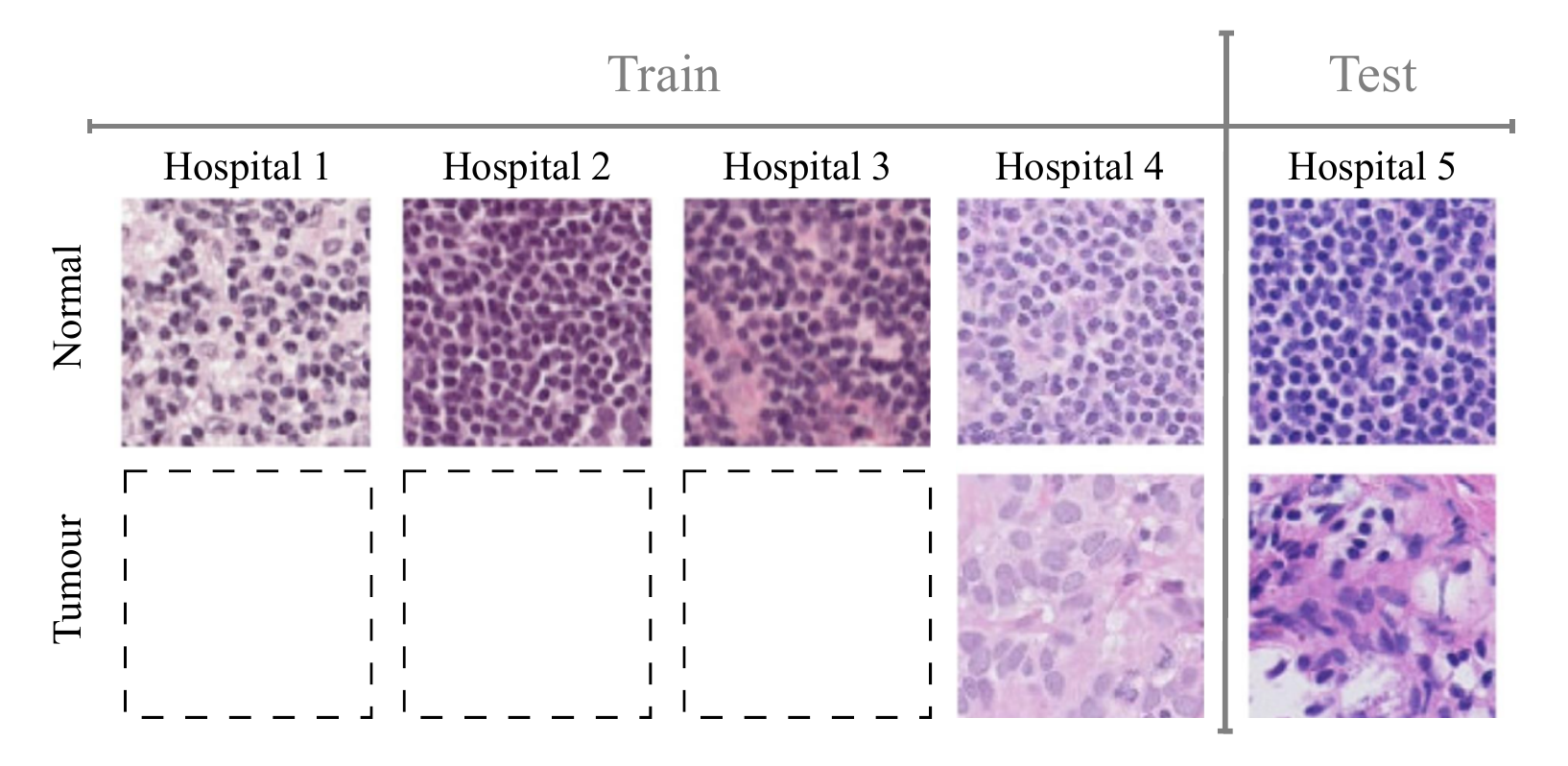}
    \caption{\redit{\textbf{Domain-linked tumour classification.} The figure illustrates the domain-linked setting on the Camelyon17 \cite{wilds-camelyon17} WILDS \cite{koh2021wilds} dataset. In this setting, models are trained on tissue images from multiple hospital domains, where tumor cases are present in only one hospital. The goal is to deploy these models more broadly across other hospitals, highlighting the challenge of generalizing to unseen clinical environments.}}
    \label{fig:data-camelyon17}
    \vspace{-5pt}
\end{figure}

\begin{table*}[b]
\centering
\caption{\textbf{Visualizing the evaluation dataset characteristics.} This table describes the multi/binary-classification evaluation datasets and shows the number of $\noc$ and $\oc$ classes for each shared-class distribution setting defined in Section~\ref{sec:class-settings}.}
\label{table:table-stats}
\redit{
\resizebox{0.9\textwidth}{!}{
\begin{tabular}{|c|c|cccc|c|c|}
    \hline
    \multirow{2}{*}{\textbf{Datasets}} & \multirow{2}{*}{\textbf{Task}} & \multicolumn{4}{c|}{\textbf{Properties}} & \multicolumn{2}{c|}{\textbf{Shared Class Setting} ($|\oc|:|\noc|$)}\tstrut\Bstrut\\ \cline{3-8} 
     &  & \# Classes & \# Samples & \# Domains & Domain Variation & ~~~~~~~Low~~~~~~~ & High \Tstrut\Bstrut\\ \hline
    PACS\cite{Li2017DeeperBA_pacs} & Multiclass & 7 & 9,991 & 4 & Style-Based & 3:4 & 5:2 \Tstrut\\
    VLCS\cite{Fang2013UnbiasedML_vlcs} & Multiclass & 5 & 10,729 & 4 & Real-World & 2:3 & 4:1 \\
    OfficeHome\cite{OfficeHome} & Multiclass & 65 & 15,588 & 4 & Style-Based & 25:40 & 50:15 \\
    \hline
    Camelyon17\cite{wilds-camelyon17} & Binary-class & 2 & 455,954 & 5 & Medical & \multicolumn{2}{c|}{1:1}\tstrut\Bstrut \\ \hline
\end{tabular}}}
\end{table*}

\noindent We specifically focus on analyzing domain-linked class performance under varying (1) shared-class distribution settings (Section~\ref{sec:class-settings}), (2) inter-domain variations, and (3) number of classes. For rigorous, fair and reproducible evaluation, we mirrored the DomainBed \cite{domainbed} test-bed to be consistent with DG literature. We report class-averaged classification accuracy for $\noc$ classes. Reported values for each dataset arise from three Monte-Carlo runs for each dataset and domain with five random hyper-parameter selections (including defaults), across the $\low$ and $\high$-shared settings. Overall, our extensive experiments summarize the results of over 360 runs per baseline. 

\subsubsection{Defining Shared-Class Distribution Settings} \label{sec:class-settings} We define two shared-class distribution settings -- \emph{Low} and \emph{High} -- denoting the relative number of shared classes $|\oc|$ with respect to the total $|\tc|$; refer to Table \ref{table:table-stats}. In the \emph{Low} setting $\sim\nicefrac{1}{3}$ of the classes are domain-shared; $\sim\nicefrac{2}{3}$ in the \emph{High} setting. $\noc$ and $\oc$ classes were randomly selected and assigned round-robin to each source-domain. Domain-linked classes only exist in one training (source) domain. Domain-shared classes do not exist in all training domains; they are randomly placed in only two of the three training domains.

\subsubsection{Datasets}\label{sec:datasets} 
We evaluate {\mymethod} and all baselines on the DG literature gold standard multiclass datasets \cite{dgTPAMIsurvey} PACS\cite{Li2017DeeperBA_pacs}, VLCS\cite{Fang2013UnbiasedML_vlcs}, and OfficeHome\cite{OfficeHome}; refer to Fig.~\ref{fig:data-multiclass}. Furthermore, we analyze domain-linked performance on a more practical medical imaging task using the large-scale WILDS\cite{koh2021wilds} Camelyon17\cite{wilds-camelyon17} tumour detection dataset; refer to Fig.~\ref{fig:data-camelyon17}. Each dataset represents different inter-domain variations and class counts, robustly examining generalization capabilities; Table~\ref{table:table-stats} provides a detailed overview of each dataset.

\subsubsection{Baselines} Selected baselines cover a variety of methodologies that are well represented in DG literature and have been DomainBed benchmarked \cite{domainbed}. Therefore, we evaluate: naive empirical risk minimization, \texttt{ERM}; popular distribution-alignment, \texttt{CORAL} \cite{CORAL}; contrastive mixing, \texttt{RSC} \cite{Huang2020RCSSelfChallengingIC} predecessor, \texttt{SelfReg} \cite{Kim2021SelfRegSC}; contrastive CLIP-based \cite{clip} method, \texttt{CAD} \cite{CAD}; popular meta-learning baselines, \texttt{ARM} \cite{ARM} and \texttt{MLDG} \cite{MLDG}; adversarial meta-learning network, \texttt{Transfer} \cite{Transfer}; mutual information regularizer, \texttt{MIRO} \cite{miro} with the \texttt{SWAD} \cite{swad} optimizer; causal representations, \texttt{CausIRL}\cite{causirl}; gradient optimization, \texttt{PGrad} \cite{wang2023pgrad}; probabilistic framework, \texttt{EQRM} \cite{eastwood2022probableEQRM}; disentanglement, \texttt{ADRMX} \cite{adrmx}; uniform learning, \texttt{URM} \cite{urm}; risk distribution matching \texttt{RDM} \cite{rdm}.

\begin{table*}[t]
\tiny
\centering
\caption{\textbf{Multi-class domain-linked $\noc$ class accuracies under the \emph{Low} and \emph{High} shared-class distribution settings.} {\mymethod} consistently outperforms competing methods in the \emph{High} setting by leveraging domain-shared classes to learn robust domain-invariant representations, which is critical for generalization to unseen domains. In alignment with our theoretical expectations, improvements in the \emph{Low} setting are more modest due to the limited availability of transferable information.}
\label{table:table-results}
\redit{
\resizebox{0.85\textwidth}{!}{
\begin{tabular}{|c|cc|c|cc|c|cc|c|c|}
\hline
  \multicolumn{1}{|c|}{\multirow{2}{*}{\textbf{Algorithm}}} &
  \multicolumn{3}{c|}{\textbf{VLCS}} \Tstrut \Bstrut &
  \multicolumn{3}{c|}{\textbf{PACS}} &
  \multicolumn{3}{c|}{\textbf{OfficeHome}} &
  \multicolumn{1}{c|}{\multirow{2}{*}{\textbf{Average}}}
   \\ \cline{2-10}
  &
  Low &
  High &
  Average &
  Low &
  High &
  Average &
  Low &
  High &
  Average &
  \Tstrut \Bstrut \\ \hline
ERM &
  {\ul 50.7} &
  {\ul 51.8} &
  {\ul 51.3} &
  36.5 &
  14.7 &
  25.6 &
  38.5 &
  37.5 &
  38.0 &
  {\ul 38.28}\Tstrut \\
CORAL \cite{CORAL} &
  45.5 &
  49.8 &
  47.6 &
  33.3 &
  13.7 &
  23.5 &
  {\ul 40.7} &
  38.9 &
  39.8 &
  37.0 \\
MLDG \cite{MLDG} &
  \cellcolor[HTML]{EFEFEF}\textbf{50.8} &
  45.2 &
  48.0 &
  \cellcolor[HTML]{EFEFEF}\textbf{38.0} &
  13.8 &
  25.9 &
  38.1 &
  37.4 &
  37.8 &
  37.22 \\
ARM \cite{ARM} &
  47.7 &
  49.0 &
  48.4 &
  36.8 &
  16.2 &
  26.5 &
  39.0 &
  38.4 &
  38.7 &
  37.85 \\
SelfReg \cite{Kim2021SelfRegSC} &
  46.6 &
  41.9 &
  44.3 &
  32.4 &
  13.4 &
  22.9 &
  40.0 &
  {\ul 39.5} &
  39.8 &
  35.63 \\
CAD \cite{CAD} &
  45.5 &
  51.7 &
  48.6 &
  33.0 &
  13.1 &
  23.1 &
  36.9 &
  36.4 &
  36.7 &
  36.10 \\
Transfer \cite{Transfer} &
  48.3 &
  48.9 &
  48.6 &
  36.4 &
  16.0 &
  26.2 &
  38.1 &
  36.8 &
  37.5 &
  37.42 \\
CausIRL \cite{causirl} &
  45.8 &
  48.9 &
  47.4 &
  33.6 &
  13.3 &
  23.5 &
  \cellcolor[HTML]{EFEFEF}\textbf{40.9} &
  39.2 &
  {\ul 40.1} &
  36.95 \\
EQRM \cite{eastwood2022probableEQRM} &
  49.1 &
  45.4 &
  47.3 &
  {\ul 37.4} &
  {\ul 17.9} &
  \cellcolor[HTML]{EFEFEF}\textbf{27.7} &
  39.9 &
  37.8 &
  38.9 &
  37.92 \\
PGrad \cite{wang2023pgrad} &
  49.0 &
  40.2 &
  44.6 &
  34.4 &
  12.6 &
  23.5 &
  39.0 &
  39.0 &
  39.0 &
  35.70\Bstrut \\ \hline
FOND &
  48.0 &
  \cellcolor[HTML]{EFEFEF}\textbf{72.1} &
  \cellcolor[HTML]{EFEFEF}\textbf{60.1} &
  35.3 &
  \cellcolor[HTML]{EFEFEF}\textbf{19.1} &
  {\ul 27.2} &
  40.3 &
  \cellcolor[HTML]{EFEFEF}\textbf{40.6} &
  \cellcolor[HTML]{EFEFEF}\textbf{40.5} &
  \cellcolor[HTML]{EFEFEF}\textbf{42.57}\Tstrut\Bstrut \\
  \hline
\end{tabular}}}
\end{table*}

\begin{figure*}[t]
    \centering


  
    \begin{minipage}[c]{0.02\textwidth}
        \centering
        \rotatebox[origin=c]{90}{
            \begin{tabular}[c]{@{}c@{}}Domain-Linked ($\mathcal{Y}_{L}$)\\Accuracy\end{tabular}
        }
        \vspace{5mm}
    \end{minipage}
    \hspace{3mm}
    \begin{minipage}[c]{0.8\textwidth} 
        \centering
        \begin{tabular}{c c c c}
        
            \subfloat[VLCS]{\includegraphics[width=0.23\textwidth]{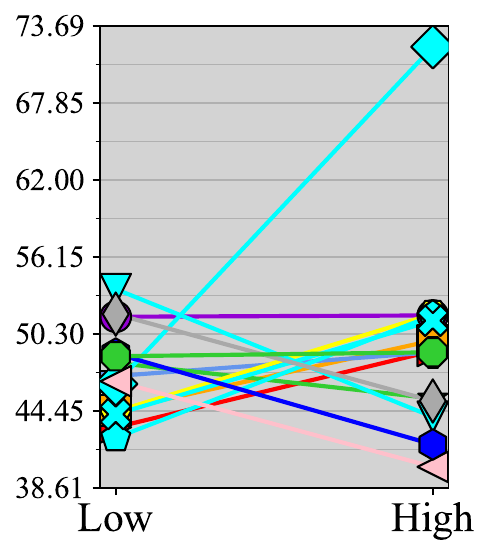}\label{fig:track_vlcs_plot}} &

            \subfloat[PACS]{\includegraphics[width=0.23\textwidth]{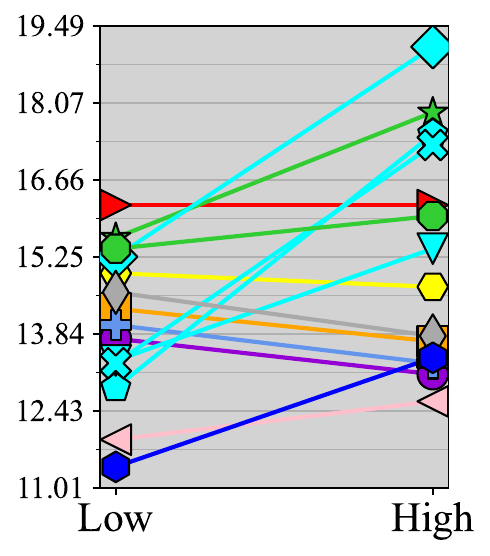}\label{fig:track_pacs_plot}} &

            \subfloat[OfficeHome]{\includegraphics[width=0.23\textwidth]{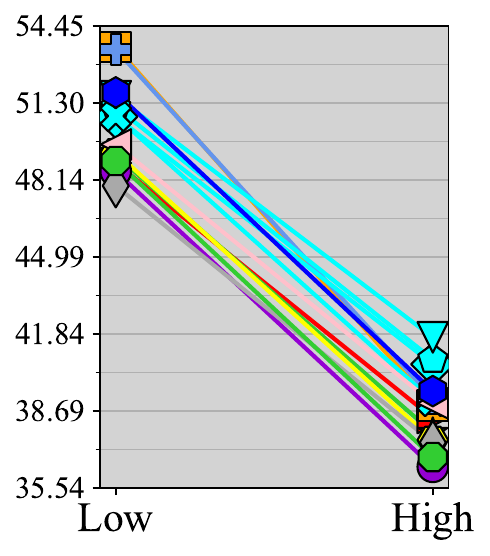}\label{fig:track_officehome_plot}} &

            \subfloat[Legend]{\includegraphics[width=0.23\textwidth]{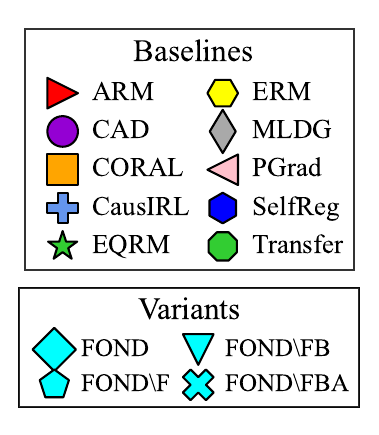}\label{fig:track_legend}} \\
            
        \end{tabular}
    \end{minipage}
    
    
    \caption{\textbf{Tracking only domain-linked $\noc$ classes present in both {\low} and {\high} shared-class distribution settings.} The key takeaway is that improvements in domain-linked class generalization do not merely result from observing more similar domain-shared classes. Otherwise, all algorithms would yield improved $\noc$ accuracies in the {\high} setting; refer to Section~\ref{sec:impact-shared-classes}.}
    \label{fig:figure-track}
\end{figure*}


\subsubsection{Implementation} For consistency, all algorithms have a fine-tuned ResNet-18 backbone \cite{he2016deep} pre-trained on ImageNet \cite{Deng2009ImageNetAL}. In alignment with DomainBed \cite{domainbed} minibatches contain an equal number of samples from each training domain. Standard DomainBed \cite{domainbed} data augmentation methods (random crops, flips, color jitter, grayscale) are applied during training. Our experiments ran on NVIDIA RTXA6000 GPUs.



\subsubsection{Hyper-Parameter Search} 
We perform hyperparameter tuning for each method via five random searches over predefined ranges stored in the DomainBed \cite{domainbed} code repository. The best hyperparameters are selected based on domain-linked class accuracy on a validation split; 80\% of the training domain data is used for training, and the other 20\% is the validation set for hyper-parameter/model selection. This maintains the out-of-distribution nature of the evaluation.

\begin{figure*}[t]
    \centering
    
    \subfloat[\label{fig:plot-radar-vlcs}][VLCS]{\includegraphics[width=0.3\textwidth]{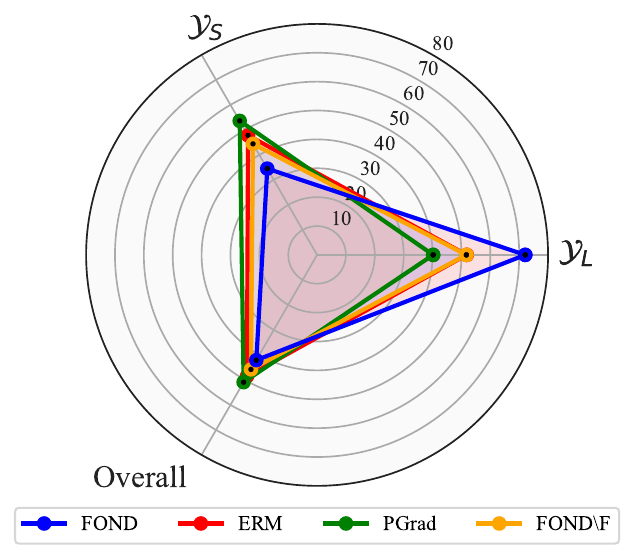}}
    \hfill
    \subfloat[ \label{fig:plot-radar-pacs}][PACS]{\includegraphics[width=0.3\textwidth]{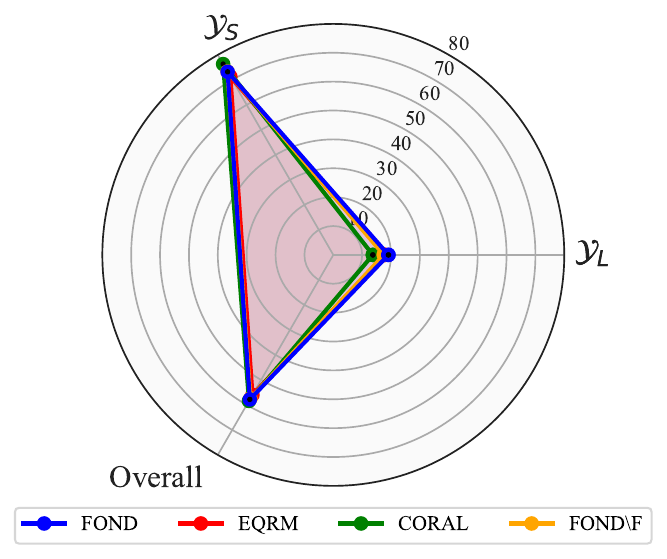}}
    \hfill
    \subfloat[\label{fig:plot-radar-officehome}][OfficeHome]{\includegraphics[width=0.3\textwidth]{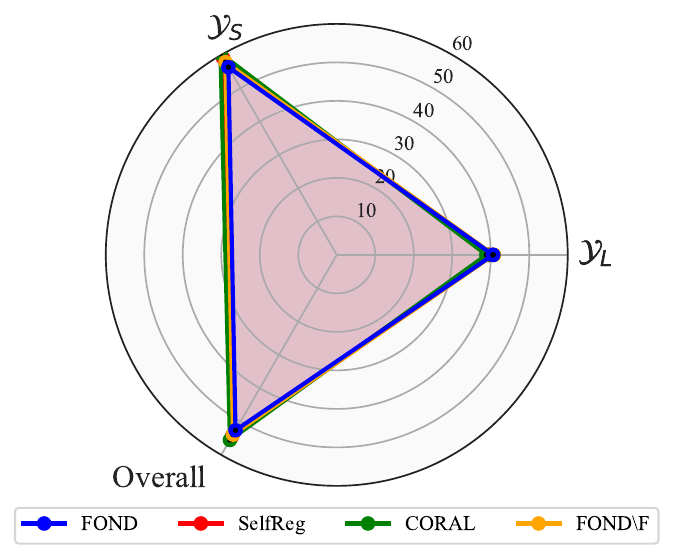}}

    \caption{\textbf{Visualizing the $\noc$ and $\oc$ performance trade-offs on the {\high} shared-class distribution setting.} We observe that {\mymethod} outperforms all baselines on $\noc$ classes while retaining competitive $\oc$ class performance, especially as the total number of target classes increases (left to right). Plotted are the best domain-linked and overall accuracy baselines.}
    \label{fig:plot-radar}
    \vspace{-10pt}
\end{figure*}

\section{Results} \label{sec:results}

\subsection{Multi-Class Image Classification}\label{sec:results-multiclass}

\subsubsection{Impact of Shared-Class Settings and Total Class Size} \label{sec:impact-shared-classes}

\redit{Table~\ref{table:table-results} provides detailed domain-linked ($\noc$) classification accuracy results for the {\low} and {\high} domain-shared class settings described in Section~\ref{sec:class-settings}. {\mymethod} \textit{consistently} outperforms all baselines in the {\high} setting by leveraging the available domain-shared class data. Strikingly, {\mymethod} results in a $39\%$ performance improvement over the best baseline (\texttt{ERM}) for VLCS. 
In the {\low} setting, {\mymethod}'s performance remains competitive despite the limited availability of transferable cross-domain information. Nonetheless, {\mymethod} consistently achieves the highest average performance across all datasets, providing a strong indication of its reliability against varying shared-class distributions.
Furthermore, note that Table~\ref{table:table-results} reports accuracies for all domain-linked classes in the {\low-\high} settings. However, in Fig.~\ref{fig:figure-track} we only track the domain-linked classes that were present in both {\low} and {\high} settings. This leads to a very important observation, namely \textit{improvements in domain-linked class generalization do not merely result from algorithms observing more similar domain-shared classes.}
Instead, the observed improvements reflect {\mymethod}'s unique capacity to explicitly transfer domain invariances learned from domain-shared classes to domain-linked ones.
Specifically, Fig.~\ref{fig:figure-track} illustrates this phenomenon for the PACS \cite{Li2017DeeperBA_pacs} and VLCS \cite{Fang2013UnbiasedML_vlcs} datasets. Except for \texttt{Transfer}\cite{Transfer}, {\mymethod} is the only algorithm that consistently improved domain-linked class accuracy from the {\low} to {\high} setting, demonstrating its targeted effectiveness in learning transferable features from domain-shared classes.
Interestingly, on the OfficeHome \cite{OfficeHome} dataset, a saturation effect emerges: adding too many domain-shared classes does not improve domain-linked performance for \textit{any} DG algorithm. We note that for small class-size datasets (PACS\cite{Li2017DeeperBA_pacs} \& VLCS\cite{Fang2013UnbiasedML_vlcs}), the transition from {\low} to {\high} entails making only 2 domain-linked classes domain-shared. However, for OfficeHome\cite{OfficeHome}, it's 25 classes, biasing the learning away from domain-linked classes.}

\subsubsection{Impact on Domain-Shared Classes} \label{sec:impact-tradeoff}
While this work aims to improve domain-linked performance, a natural question arises regarding the impact on the domain-shared classes. To analyze this performance, in Fig.~\ref{fig:plot-radar}, we present a comparative analysis of the $\noc$ vs. $\oc$ vs. overall accuracy. For each dataset, {\mymethod} results in a striking improvement in $\noc$ accuracy compared to the best baselines while retaining competitive $\oc$ and overall accuracy. We also observe that different {\mymethod} variants ({\variantF}) can be used to reconcile the trade-off in the real world.

\begin{table*}[t]
    \tiny
    \caption{\textbf{Domain-linked tumour classification accuracy on the Camelyon17\cite{wilds-camelyon17} WILDS\cite{koh2021wilds} dataset.} Algorithms are trained on tissue images from four hospital domains—only one of which includes positive tumour cases—and evaluated on a held-out fifth hospital. The training and evaluation splits/domains adhere to the WILDS \cite{koh2021wilds} protocol to ensure unbiased assessment. Refer to Section~\ref{sec:results-camelyon17} and Fig.~\ref{fig:data-camelyon17}. The "Upper-Bound" reflects if all hospital data had positive tumour cases.}
    \label{table:camelyon17}
    \centering
    \redit{
    \resizebox{0.7\textwidth}{!}{
    \begin{tabular}{|c|cc|c|cc|c|}
    \hline
    \multicolumn{1}{|c|}{\multirow{2}{*}{\textbf{Algorithm}}} & \multicolumn{3}{c|}{In-Distribution (Validation Set)} & \multicolumn{3}{c|}{\textbf{Out-of-Distribution (Test Set)}}\Tstrut\Bstrut \\ \cline{2-7} 
     & Normal & Tumour & Average & \textbf{Normal} & \textbf{Tumour} & \textbf{Average}\Tstrut\Bstrut \\ \hline
    ERM (Upper-Bound) & 98.4 & 96.8 & 97.6 & 99.6 & 80.6 & 94.6\Tstrut \\
    \hline
    ERM & 98.8 & 96.2 & {97.5} & 99.9 & 49.6 & 74.8\Tstrut \\
    CORAL \cite{CORAL} & 99.1 & {96.5} & 97.8 & 99.9 & 54.9 & 77.4 \\
    MLDG \cite{MLDG} & {99.2} & 95.3 & 97.2 & 100.0 & 55.6 & 77.8 \\
    ARM \cite{ARM} & 98.4 & 94.3 & 96.4 & 100.0 & 48.8 & 74.4 \\
    SelfReg \cite{Kim2021SelfRegSC} & 99.2 & 95.7 & 97.4 & 99.9 & 51.8 & 75.8 \\
    Transfer \cite{Transfer} & 98.6 & 94.1 & 96.3 & 99.9 & {\ul 58.7} & {\ul 79.3} \\
    CAD \cite{CAD} & 99.2 & 90.4 & 94.8 & 100.0 & 28.3 & 64.1 \\
    CausIRL \cite{causirl} & 98.6 & 96.0 & 97.3 & 100.0 & 29.2 & 64.6 \\
    EQRM \cite{eastwood2022probableEQRM} & 98.5 & 96.4 & 97.4 & 100.0 & 45.7 & 72.9 \\
    MIRO \cite{miro} & 98.4 & 95.1 & 96.7 & 100.0 & 22.4 & 61.2 \\
    MIRO\cite{miro} + SWAD\cite{swad} & 97.8 & 95.2 & 96.5 & 100.0 & 27.8 & 63.9 \\
    PGrad \cite{wang2023pgrad} & 97.1 & 95.3 & 96.2 & 99.9 & 56.6 & 78.3 \\
    ADRMX \cite{adrmx} & 98.1 & 91.9 & 95.0 & 100 & 58.4 & 79.2 \\
    RDM \cite{rdm} & 99.2 & 97.0 & 98.1 & 99.9 & 43.4 & 71.6  \\
    URM \cite{urm} & 98.7 & 94.8 & 96.7 & 100.0 & 35.2 & 67.6 \\
    \hline
    FOND & 96.7 & 96.7 & 96.7 & 99.8 & \cellcolor[HTML]{EFEFEF}\textbf{67.0} & \cellcolor[HTML]{EFEFEF}\textbf{83.4}\Tstrut\Bstrut \\ \hline
    \end{tabular}}}
\end{table*}

\begin{table}[t]
\centering
\caption{\textbf{Ablation study on {FOND}'s components.} This table reports average multiclass domain-linked $\noc$ accuracies.}
\label{table:table-ablations}
\redit{
\resizebox{0.48\textwidth}{!}{
\begin{tabular}{|c|cccc|ccc|c|}
\hline
\multirow{2}{*}{\textbf{Variant}} & \multicolumn{4}{c|}{\textbf{Components}} & \multicolumn{3}{c|}{\textbf{Datasets} (Average Acc.)} & \multirow{2}{*}{\textbf{Average}}\Tstrut\Bstrut \\ \cline{2-8} 
 & $\mathcal{L}_{x}$ & $\alpha$ & $\beta$ & $\mathcal{L}_{d}$ & VLCS & PACS & OfficeH. & \Tstrut\Bstrut \\ \hline
FOND\textbackslash{}FBA & $\checkmark$ &  &  &  & 49.0 & 26.4 & 40.1 & 39.0\Tstrut \\
FOND\textbackslash{}FB & $\checkmark$ & $\checkmark$ &  &  & 47.0 & 24.3 & \cellcolor[HTML]{EFEFEF}\textbf{41.4} & 37.6\\
FOND\textbackslash{}F & $\checkmark$ & $\checkmark$ & $\checkmark$ &  & {\ul 50.1} & {\ul 26.4} & {\ul 40.0} & {\ul 39.1}\Bstrut \\
\hline
FOND & $\checkmark$ & $\checkmark$ & $\checkmark$ & $\checkmark$ & \cellcolor[HTML]{EFEFEF}\textbf{60.1} & \cellcolor[HTML]{EFEFEF}\textbf{27.20} & 40.5 & \cellcolor[HTML]{EFEFEF}\textbf{42.6} \Tstrut\Bstrut \\ \hline
\end{tabular}}}
\end{table}

\subsubsection{Impact of Domain Variation Types} \label{sec:impact-variations}
\redit{The types of variations across domains significantly influence domain-linked class performance. To examine this impact, we compare the VLCS \cite{Fang2013UnbiasedML_vlcs} and PACS \cite{Li2017DeeperBA_pacs} datasets, which are comparable in terms of class count and dataset size but differ distinctly in the nature of their domain variations; refer to Table~\ref{table:table-stats} and Fig.~\ref{fig:data-multiclass}.
The VLCS dataset exemplifies subtle, real-world variations, such as differences in camera angles, illumination, and environmental contexts. Conversely, PACS contains more prominent, style-based differences, including significant variations in textures, artistic styles, and visual appearances. According to Theorem~\ref{thm:main}, generalization errors for domain-linked classes are closely tied to the distribution divergence between source and target domains: $d_{\mathcal{H}\Delta\mathcal{H}}(T, S^{\ell})$. Thus, greater divergence is expected to deteriorate model performance.
This theoretical insight aligns with empirical observations illustrated in Fig.~\ref{fig:plot-radar}. Models evaluated on PACS, consistently yield lower accuracy scores for domain-linked classes compared to VLCS. Additionally, PACS exhibits a larger discrepancy between the performances of domain-linked $\noc$ and domain-shared $\oc$ classes.}

\subsubsection{Visualizing Learned Representations} \label{sec:tsne}
\begin{figure*}[!t]

    \resizebox{.9\textwidth}{!}{
    \hspace{-150pt}
    \centering
    \begin{minipage}[c]{0.41\textwidth} 
        \centering
        \hspace{-10pt}
        \rotatebox[origin=c]{90}{\includegraphics[width=\textwidth]{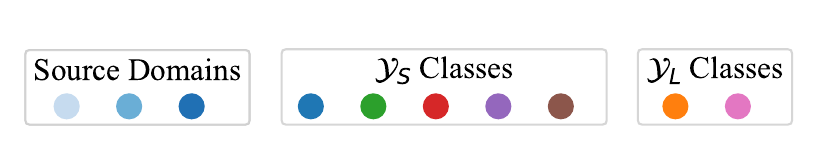}}
        \hspace{-150pt}
    \end{minipage}%
    \begin{minipage}[c]{0.98\textwidth} 
        \centering
        \begin{tabular}{c c c}
        
            \subfloat[ERM source data (classes)]{\includegraphics[width=0.3\textwidth]{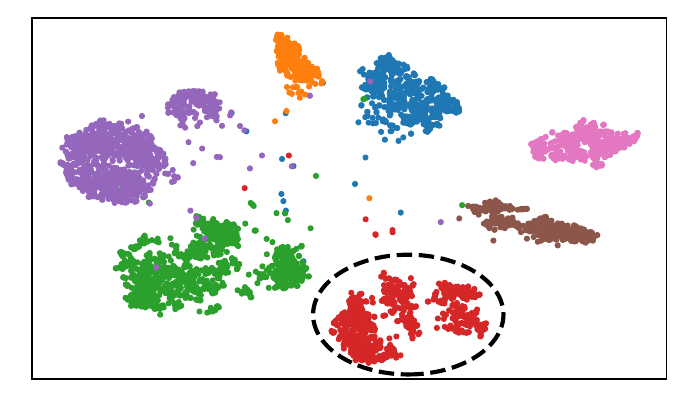}\label{fig:tsne_pacs_erm_source}} &
            
            \subfloat[ERM source data (domains)]{\includegraphics[width=0.3\textwidth]{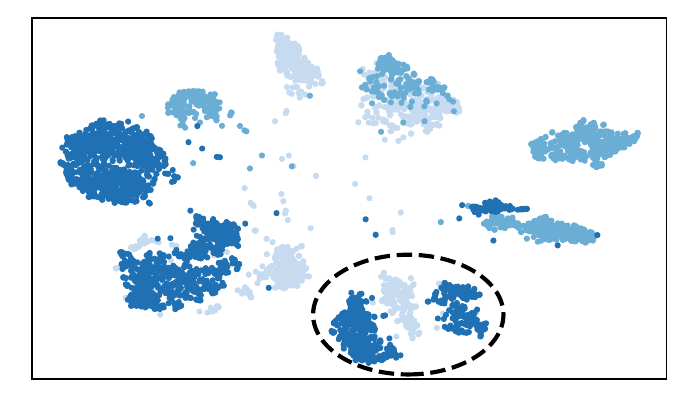}\label{fig:tsne_pacs_erm_domain}} &
            
            \subfloat[ERM target data (classes)]{\includegraphics[width=0.3\textwidth]{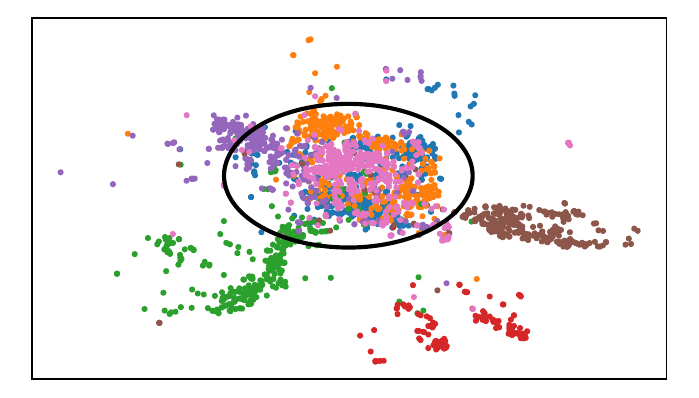}\label{fig:tsne_pacs_erm_target}} \\
            
        \end{tabular}
        \begin{tabular}{c c c}
        
            \subfloat[ARM source data (classes)]{\includegraphics[width=0.3\textwidth]{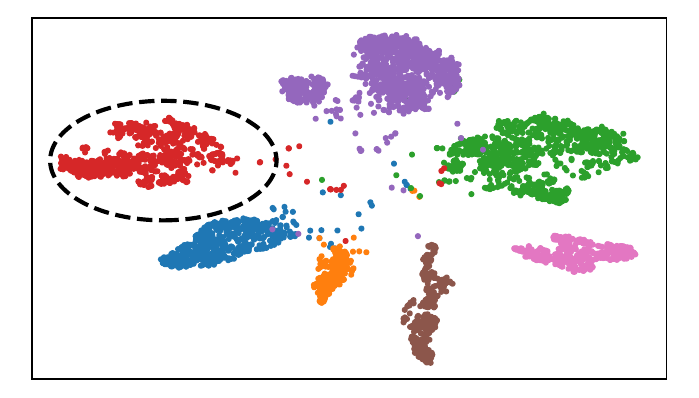}\label{fig:tsne_pacs_arm_source}} &
            
            \subfloat[ARM source data (domains)]{\includegraphics[width=0.3\textwidth]{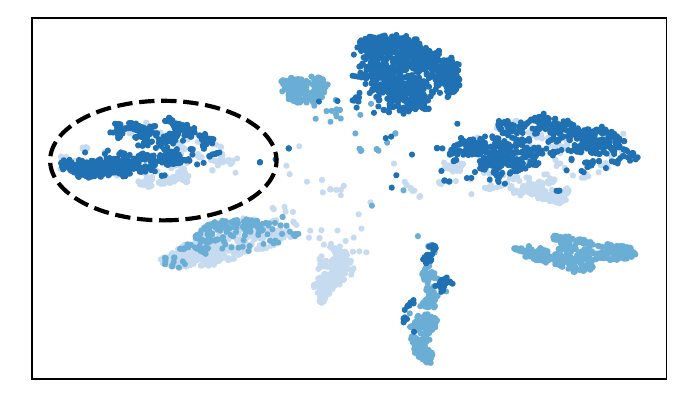}\label{fig:tsne_pacs_arm_domain}} &
            
            \subfloat[ARM target data (classes)]{\includegraphics[width=0.3\textwidth]{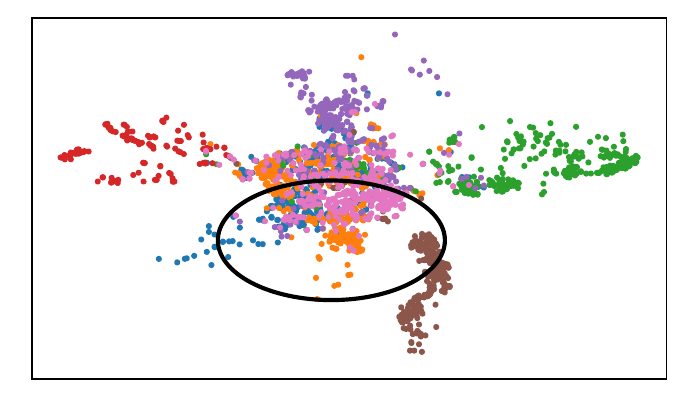}\label{fig:tsne_pacs_arm_target}} \\
            
        \end{tabular}
        \begin{tabular}{c c c}

            \subfloat[{\MYmethod} source data (classes)]{\includegraphics[width=0.3\textwidth]{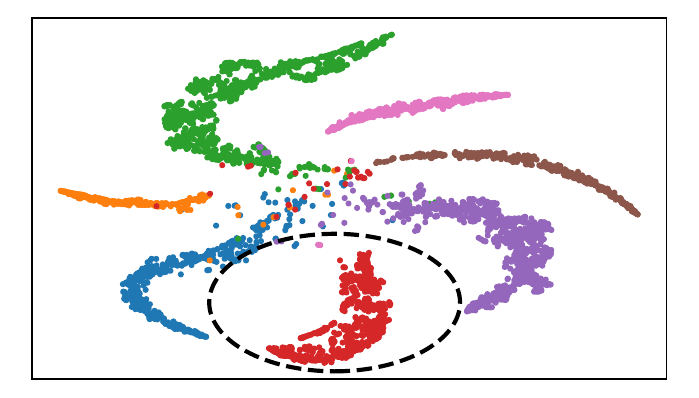}\label{fig:tsne_pacs_ours_source}} &
            
            \subfloat[{\MYmethod} source data (domains)]{\includegraphics[width=0.3\textwidth]{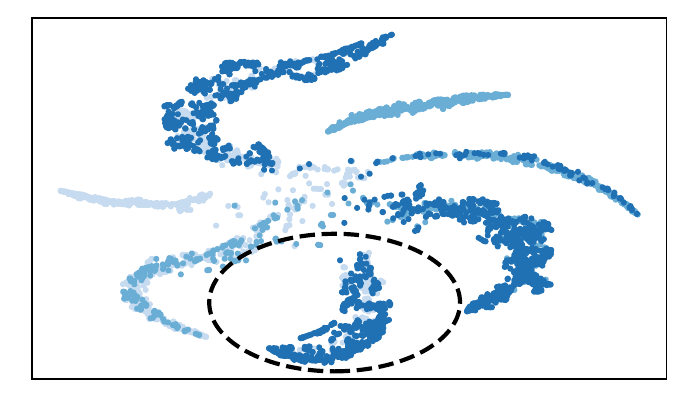}\label{fig:tsne_pacs_ours_domain}} &
            
            \subfloat[{\MYmethod} target data (classes)]{\includegraphics[width=0.3\textwidth]{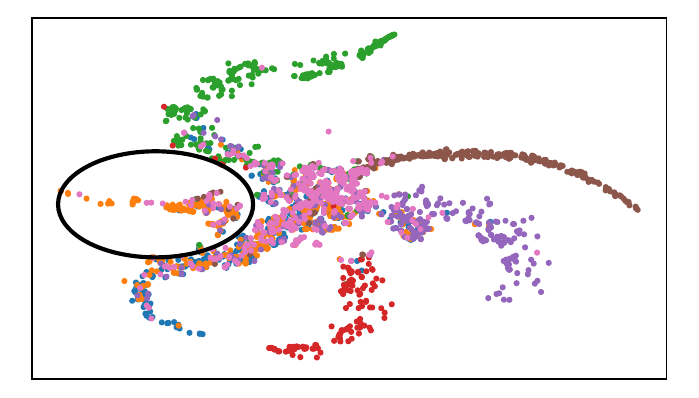}\label{fig:tsne_pacs_ours_target}} \\
        \end{tabular}
    \end{minipage}}
    \caption{\textbf{t-SNE latent $\mathbf{h}$ representation visualization for the PACS-{\high} dataset}. Each row visualizes the representations of the naive (\texttt{ERM}), top-performing-baseline (\texttt{ARM}) and our ({\mymethod}) algorithm. Source-domain (\emph{Photo}, \emph{Art} and \emph{Sketch}) representations are colored by class and domain. Target-domain (\emph{Cartoon}) representations are colored by class. We highlight domain-linked $\noc$ class generalization (solid circle) and domain-invariant learning (broken circle) in Section~\ref{sec:tsne}.}
    \label{fig:tsne_plots}
    \vspace{-5pt}
\end{figure*}

In Fig.~\ref{fig:tsne_plots} we visualize latent representations from the PACS-\emph{High} dataset via t-SNE plots \cite{tsne}. To gain insight into our method's strong performance during the \emph{High} shared-class distribution setting, we analyze the representations learned by \texttt{ERM} (naive-baseline), \texttt{ARM} (top-performing-baseline) and {\mymethod} (top-performing-method); refer to Table~\ref{table:table-results}. On source-domain data, the \texttt{ERM} class-colored clusters are distinctly sub-clustered by domain (e.g. broken circle in Fig.~\ref{fig:tsne_pacs_erm_source} and \subref{fig:tsne_pacs_erm_domain}). Whereas \texttt{ARM} (Fig.~\ref{fig:tsne_pacs_arm_source} and \subref{fig:tsne_pacs_arm_domain}) and {\mymethod} (Fig.~\ref{fig:tsne_pacs_ours_source} and \subref{fig:tsne_pacs_ours_domain}) demonstrate more domain-invariant representations since their classes are not as distinctly clustered by domain. Furthermore, for domain-linked $\noc$ class samples, {\mymethod} (e.g. solid circle in Fig.~\ref{fig:tsne_pacs_ours_target}) yields more generalizable representations than \texttt{ARM} (Fig.~\ref{fig:tsne_pacs_arm_target}) and \texttt{ERM} (Fig.~\ref{fig:tsne_pacs_erm_target}). While all methods struggle on the pink class, {\mymethod} empirically maintains top performance.


\subsection{Binary Domain-Linked Tumour Classification} \label{sec:results-camelyon17}
\subsubsection{Motivation} \redit{A prominent challenge in medical imaging is the scarcity of labeled data, as obtaining expert annotations (e.g., from physicians or radiologists) is costly, labor-intensive, and time-consuming\cite{ZHANG2024108238}. This challenge often forces medical imaging models to be trained on labeled data aggregated from multiple distinct sources (e.g., hospitals), aiming for broad deployment. However, variations in data collection protocols and processing techniques across institutions significantly degrade model accuracy when deployed in new clinical environments \cite{zech2018variable}. Specifically, in histopathology, variations arise from patient demographics, staining procedures, and imaging acquisition processes \cite{tellez2019quantifying_hist}. Additionally, initial collections for novel medical imaging tasks, such as identifying Covid-19 from chest X-rays \cite{wang2020covid}, often originate from a single laboratory, resulting in new diagnostic classes being domain-linked.}

\subsubsection{Setup} \redit{To simulate this practical scenario, we use the WILDS\cite{koh2021wilds} Camelyon17 \cite{wilds-camelyon17} tumour classification dataset, comprising approximately 500,000 histopathology images across five hospital domains. The original binary classification task distinguishes between tumourous and normal tissues, training on data from the first four hospitals and evaluating on a held-out fifth. To replicate the conditions typical in early-stage diagnostic model development, we restrict positive tumour cases exclusively to the fourth hospital, making this class domain-linked (Fig.~\ref{fig:data-camelyon17}). This arrangement mirrors real-world conditions wherein acquiring diverse samples of normal tissue is relatively straightforward, whereas early positive tumour cases typically originate from limited sources.}

\subsubsection{Results} \redit{ Table~\ref{table:camelyon17} presents tumour (domain-linked) and normal tissue (domain-shared) detection accuracies across various methods, evaluated on an in-distribution validation (20\% of each training hospital is held out) and out-of-distribution test (held-out fifth hospital) dataset. While all methods show comparable and strong detection accuracies on the in-distribution validation set, {\mymethod} significantly outperforms the best-performing baseline, \texttt{Transfer}, by nearly 9\% in detecting tumours on the out-of-distribution test data. The marked performance gap between in-distribution and out-of-distribution data highlights FOND’s superior robustness in representing tumour data, underscoring its practical advantage in domain-linked medical classification scenarios.}

\subsection{Ablation Study}\label{sec:ablations} \redit{
\noindent Table~\ref{table:table-ablations} evaluates each of \texttt{FOND}'s variants/components in the order they were introduced in Section~\ref{sec:methodology}. The table reports the average for the {\low} and {\high} settings. 
Beginning with the simplest variant, FOND\textbackslash{}FBA, which incorporates only the basic domain-invariant learning objective $\mathcal{L}_{xdom}$, we observe an average accuracy of 39.0\%. Incorporating inter-domain positive weighting through the $\alpha$ parameter (FOND\textbackslash{}FB) yields the best performance on in scenarios with increased class counts, i.e., OfficeHome with $\sim$10$\times$ more classes. As the number of classes increases, the diversity of positive inter-domain pairs also increases, making the $\alpha$-only variant FOND\textbackslash{}FB the strongest in such settings (including all baselines in Table~\ref{table:table-results}) despite its overall average accuracy of 37.6\%.
Introducing the intra-domain negative weighting via the $\beta$ parameter (FOND\textbackslash{}F) balances the trade-offs of the $\alpha$-only variant FOND\textbackslash{}FB and outperforms all other baselines on average in Table~\ref{table:table-results}, second only to FOND.
Finally, integrating the fairness-inspired discrepancy loss $\mathcal{L}_{disc}$ alongside all previous components culminates in the complete FOND model, which achieves the highest average accuracy of 42.6\%. These results validate the synergistic effectiveness of the contrastive and fairness-inspired components in the proposed methodology.
}

\section{Discussion} \label{sec:discussion}

\noindent \textbf{Summary.} This work introduced the domain-linked domain generalization (DG) task, highlighting a critical challenge in machine learning in which specific classes appear exclusively within certain domains due to data scarcity and collection constraints. Conventional DG efforts focus primarily on improving overall accuracy, which we observe can lead to critical failures in the real world. We proposed {\mymethod}, a fairness-inspired contrastive learning framework, designed explicitly to leverage knowledge from domain-shared classes to improve generalization for domain-linked classes. Our extensive empirical evaluations demonstrated that {\mymethod} achieves state-of-the-art performance, outperforming existing DG approaches, particularly in realistic, clinically relevant scenarios.


\vspace{2pt}
\noindent \noindent\textbf{Significance.} Introducing the domain-linked DG task provides a realistic and challenging benchmark that reflects real-world data collection limitations and spurs meaningful methodological developments in the DG community. By identifying severe performance discrepancies between domain-linked and domain-shared classes, this research contributes vital insights into the limitations of current DG methodologies, motivating further exploration and innovation in this area.


\vspace{2pt}
\noindent \noindent\textbf{Key takeaways.} Our results reveal that domain-linked classes significantly underperform under standard DG setups, even when generalization improves for domain-shared classes. FOND effectively mitigates this gap by leveraging contrastive and fairness-inspired objectives to transfer domain-invariant representations. Notably, improvements in domain-linked class accuracy do not merely stem from the presence of more similar classes but from FOND’s ability to explicitly utilize domain-shared knowledge. Moreover, it achieves these gains with minimal trade-offs in shared-class or overall performance. Furthermore, {\mymethod} was the only method to achieve reliable performance under the nuanced real-world domain variations found in the VLCS and medical Camelyon17 dataset. These insights offer practical guidance for handling data scarcity and class imbalance in real-world DG tasks.


\vspace{2pt}
\noindent\textbf{Limitations and future work.} Like all baselines, {\mymethod} also struggles in the $\low$-shared setting due to the reduction in shared classes to learn from. Next, we make a binary distinction between domain-linked and domain-shared classes, but exploring methods that can adapt to varying levels of shared-\emph{ness} in future will be exciting. Nevertheless, our work calls for paying closer attention to DG biases and assumptions.



\section*{Acknowledgments}
The work of Sirisha Rambhatla was supported by the Natural Sciences and Engineering Research Council of Canada (NSERC) Discovery Grant, under Grant RGPIN-2022-03512.

\bibliographystyle{IEEEtranN}
\bibliography{references}

\end{document}